\documentclass[journal]{IEEEtran}
\usepackage{graphicx}
\usepackage{multirow}
\usepackage{subfiles}
\usepackage{color}
\usepackage{soul}
\usepackage{amssymb}
\usepackage{amsmath}
\usepackage{url}

\newcommand{\beq}{\begin{equation}}
\newcommand{\eeq}{\end{equation}}
\newcommand{\beqn}{\begin{equation*}}
\newcommand{\eeqn}{\end{equation*}}

\title{A 3D Probabilistic Deep Learning System for Detection and Diagnosis of Lung Cancer Using Low-Dose CT Scans}
\author{Onur Ozdemir, \IEEEmembership{Member, IEEE,} Rebecca L. Russell, and Andrew A. Berlin, \IEEEmembership{Member, IEEE}
	\thanks{Copyright (c) 2019 IEEE. Personal use of this material is permitted. Permission from IEEE must be obtained for all other uses, in any current or future media, including reprinting/republishing this material for advertising or promotional purposes, creating new collective works, for resale or redistribution to servers or lists, or reuse of any copyrighted component of this work in other works.
		
	This work was supported by the Charles Stark Draper Laboratory IR\&D program under the guidance of Amy Duwel and Sheila Hemami.
		
	O. Ozdemir was with Draper, Cambridge, MA 02139 when this work was performed. He is now with Google, Cambridge, MA 02142 (email: oozdemir@syr.edu).
	
	R. L. Russell and A. A. Berlin are with Draper, Cambridge, MA, 02139 (email: rrussell@draper.com; andrew.berlin@ieee.org.)}}
\begin{document}

\maketitle

\begin{abstract}

We introduce a new computer aided detection and diagnosis system for lung cancer screening with low-dose CT scans that produces meaningful probability assessments. Our system is based entirely on 3D convolutional neural networks and achieves state-of-the-art performance for both lung nodule detection and malignancy classification tasks on the publicly available LUNA16 and Kaggle Data Science Bowl challenges.
While nodule detection systems are typically designed and optimized on their own, we find that it is important to consider the coupling between detection and diagnosis components. Exploiting this coupling allows us to develop an end-to-end system that has higher and more robust performance and eliminates the need for a nodule detection false positive reduction stage.
Furthermore, we characterize model uncertainty in our deep learning systems, a first for lung CT analysis, and show that we can use this to provide well-calibrated classification probabilities for both nodule detection and patient malignancy diagnosis. These calibrated probabilities informed by model uncertainty can be used for subsequent risk-based decision making towards diagnostic interventions or disease treatments, as we demonstrate using a probability-based patient referral strategy to further improve our results.

\end{abstract}

\section{Introduction}

Lung cancer is both one of the most common cancers and the leading cause of cancer death, accounting for approximately a quarter of all cancer related deaths in the US \cite{acs_18}. The high mortality rate associated with lung cancer is in part because its symptoms become apparent only after the cancer is already at an advanced stage. Low-dose Computerized Tomography (CT) has been proposed as a safe and effective tool for preventative screening of high-risk populations. Annual CT screening could reduce lung cancer mortality by at least $20\%$ after $7$ years relative to annual chest radiography \cite{nlst11}.

While lung CT screening has the potential to dramatically reduce the number of lung cancer related deaths, the burden on radiologists to make screening accurate and efficient for large volumes of CT scans is high. Automated algorithmic solutions may help reduce this burden, but interfacing between algorithmic solutions and doctors is also a challenge when these algorithms cannot reliably communicate their uncertainty. To address these needs, we introduce an end-to-end probabilistic diagnostic system for lung cancer built on deep 3D convolutional neural networks (CNNs). Our system directly analyzes CT scans and provides calibrated probabilistic scores that accurately characterize uncertainty. 

Our system has two main components: 1) a Computer-Aided Detection (CADe) module that detects and segments suspicious lung nodules, and 2) a Computer-Aided Diagnosis (CADx) module that performs both nodule-level assessment and patient-level malignancy classification by analyzing suspicious lesions from CADe. Both our CADe and CADx modules achieve comparable to or better performance than the best published CADe and CADx systems on the LUNA16 \cite{luna16} and Kaggle Data Science Bowl 2017 \cite{dsb} benchmarks, even though our system is only trained on the limited data and labels provided by these datasets, in contrast to \cite{kaggle17, philips18}, which utilizes additional training data. 

We further demonstrate that the results from our system have meaningful probabilistic interpretations because our approach integrates model uncertainty in all classification evaluations. To the best of our knowledge, model uncertainty has not been considered in the context of lung CT analysis before. Since model uncertainty allows us to quantify the uncertainty in model predictions emanating from lack of training data \cite{gal_icml16}, by quantifying and including model uncertainty, our system can  more reliably assess out-of-cohort data without making overly-confident predictions, making it more trustworthy for real-world usage. We show that the incorporation of model uncertainty makes the outputs of both CADe and CADx systems well-calibrated to the original data distribution and thus the probabilities output by our systems are directly interpretable as real probabilities. Based on these probabilities, we further propose patient referral strategies for both CADe and CADx.

An important aspect of any CADe/CADx system is the coupling between different processing and decision making components and how the performance of each component affects the overall performance. Malignant nodules may be missed by a CADe system tuned for high precision, while excess false positive candidates from a CADe system with high recall may overburden any CADx system that has not been developed to be resilient to those types of false positive candidates. The recall-precision trade-off of the CADe systems used both for generating training and evaluation nodule candidates directly impacts the performance of the CADx system. Due to this inherent coupling, we develop and study both components together. 

The notion of jointly-developed CADe/CADx challenges common practice in the lung screening industry, where CADe and CADx are considered as independent products. Much of the previously published research focuses on optimizing either CADe alone \cite{jesson_miccai17, duo_miccai17, ding_miccai17, ross_jamia18} or CADx alone \cite{shen_ipmi15, shen_miccai16, bagci_miccai16, bagci_isbi17, bagci_ipmi17, bagci_embc18}, without careful consideration of how they are affected by each other. 
Patient-level end-to-end diagnosis using machine learning with diagnosis-confirmed labels has been largely unaddressed until the release of the Kaggle Data Science Bowl \cite{dsb} dataset, which encourages the development of joint CADe/CADx systems that achieve the best end-to-end performance.

We for the first time systematically study the CADe/CADx coupling and show that an individually-optimized CADe system (e.g., for the LUNA16 challenge) can be highly suboptimal for automated CADx, i.e., cancer diagnosis. 
While much effort is typically put into reducing the false positives from CADe, our CADx component is not only robust to our maximum CADe false-positive rate (FPR) of approximately $8$ candidates per patient (the largest of the proposed LUNA16 evaluation points) but actually performs much better when trained on this data compared to a cleaner sample. This demonstrates that CADe approaches optimized for performance on the LUNA16 CADe benchmark may not perform as well when used as part of a full CADe/CADx system.

The rest of the paper is organized as follows. In Section \ref{sec:relwork}, we review the relevant literature on deep learning based CADe/CADx approaches to address lung cancer diagnosis and briefly compare the most relevant ones to our system. Our overall system model is explained in Section \ref{sec:sysmodel}. Section \ref{sec:data} describes the datasets we employ to train and evaluate our models. The details of the developed CADe and CADx models are provided in Sections \ref{sec:cade} and \ref{sec:cadx}, respectively. We present our CADe/CADx results and coupling analysis in Section \ref{sec:results} and provide a discussion on calibration and referral strategies in Section \ref{sec:referral}. Finally, concluding remarks are presented in Section \ref{sec:conc}.

\section{Related Work}
\label{sec:relwork}

While traditional CAD methods have relied on hand-engineered features and conventional image processing methods \cite{sluimer_tmi06}, recent advances in deep learning for computer vision have resulted in a major shift towards deep learning based solutions \cite{shin_tmi16}. Multiple independent studies have shown that deep learning outperforms traditional approaches in lung CT analysis and results in more robust solutions, thanks to its ability to automatically learn useful feature representations from data \cite{shen_ipmi15, setio_tmi16}. Therefore, while the literature on lung CT analysis is vast, we focus only on related work based on deep learning in this section.

Although earlier deep learning CADe approaches used 2D image slices \cite{setio_tmi16}, 3D volumetric images provide more informative features leading to improved performance \cite{jesson_miccai17, duo_miccai17, ding_miccai17, zhu_wacv18, bagci_miccai18, ross_jamia18}. The majority of proposed CADe approaches consist of two steps \cite{jesson_miccai17, duo_miccai17, ding_miccai17, ross_jamia18}: 1) a nodule candidate extraction step that uses a segmentation network, e.g., \cite{vnet_3dv16}, or a region proposal network, e.g., \cite{frcnn_pami17}; and 2) a false positive reduction step that uses a 3D CNN model for nodule-level detection. Most recently, the authors in \cite{bagci_miccai18} proposed a single-step CADe model that is based on a 3D dense CNN to perform nodule detection in one step. Their model achieved a LUNA16 score of $0.897$, outperforming previously proposed CADe models. In contrast, our combined CADe/CADx system does not require a second false positive reduction step since our CADx model is relatively insensitive to false positives. However, we did train a CADe scoring network in order to produce a comparable LUNA16 benchmark score and achieved a LUNA16 score of $0.921$, which, to the best of our knowledge, is the best published result on this dataset.

The majority of deep learning based lung CADx approaches use the LIDC-IDRI \cite{armato2011} dataset, which is labeled with nodule annotations and malignancy risk scores from four different radiologists \cite{shen_ipmi15, shen_miccai16, bagci_miccai16, bagci_isbi17, bagci_ipmi17, bagci_embc18, zhu_wacv18}. Although this dataset is useful for developing  nodule detection and nodule-level malignancy prediction models, the associated malignancy labels represent subjective opinions of the annotator radiologists, which have not been confirmed by pathology diagnosis. There is also significant inter-observer variability among radiologists \cite{bagci_miccai16}, which complicates the verification of obtained CADx results. Further, the vast majority of published research work addresses either CADe or CADx alone, whereas in real life the performance of the overall CADx system is directly impacted by the performance of the preceding CADe model. In addition to the LIDC-IDRI dataset, which has nodule-level annotations, a second dataset that was released by the National Cancer Institute for the 2017 Data Science Bowl on Kaggle has patient-level binary malignancy labels. The Kaggle dataset does not have nodule-level annotations, but its patient-level labels have been confirmed by pathology diagnosis. This is more representative of a real world test scenario where a CAD system gets access to raw CT scans and has to provide end-to-end malignancy decisions based only on those scans without relying on external nodule annotation information. 

There are two research papers that propose an end-to-end CADe/CADx system and use the Kaggle dataset similar to ours \cite{kaggle17, philips18}. The first one, \cite{kaggle17}, uses a 3D region-based convolutional neural network (R-CNN) for nodule detection followed by multiple instance learning using a leaky noisy-OR combination approach for malignancy classification. The authors of \cite{kaggle17} examined each CT scan in the Kaggle training data and hand-annotated suspicious nodules to curate new training data. They then used those annotations along with the LIDC-IDRI dataset to train their nodule detection, i.e., CADe, model. Their final system achieves a CADx performance of $0.87$ Area Under the ROC (AUROC) curve on the Kaggle test set, which, to the best of our knowledge, is the best published CADx performance. In contrast, our models were not trained with any additional annotations besides what have been provided by the LIDC-IDRI and Kaggle datasets, and still achieve an equally good CADx performance. The second CADe/CADx paper that uses the Kaggle dataset \cite{philips18} uses models trained on the NLST dataset \cite{nlst11}, which is a superset of the Kaggle dataset and includes almost twice as much training data as the Kaggle training data, and achieves a CADx performance of $0.84$ AUROC on the Kaggle test set.

\section{CAD System Model}
\label{sec:sysmodel}

\begin{figure}
\begin{center}
\includegraphics[width=0.45\textwidth]{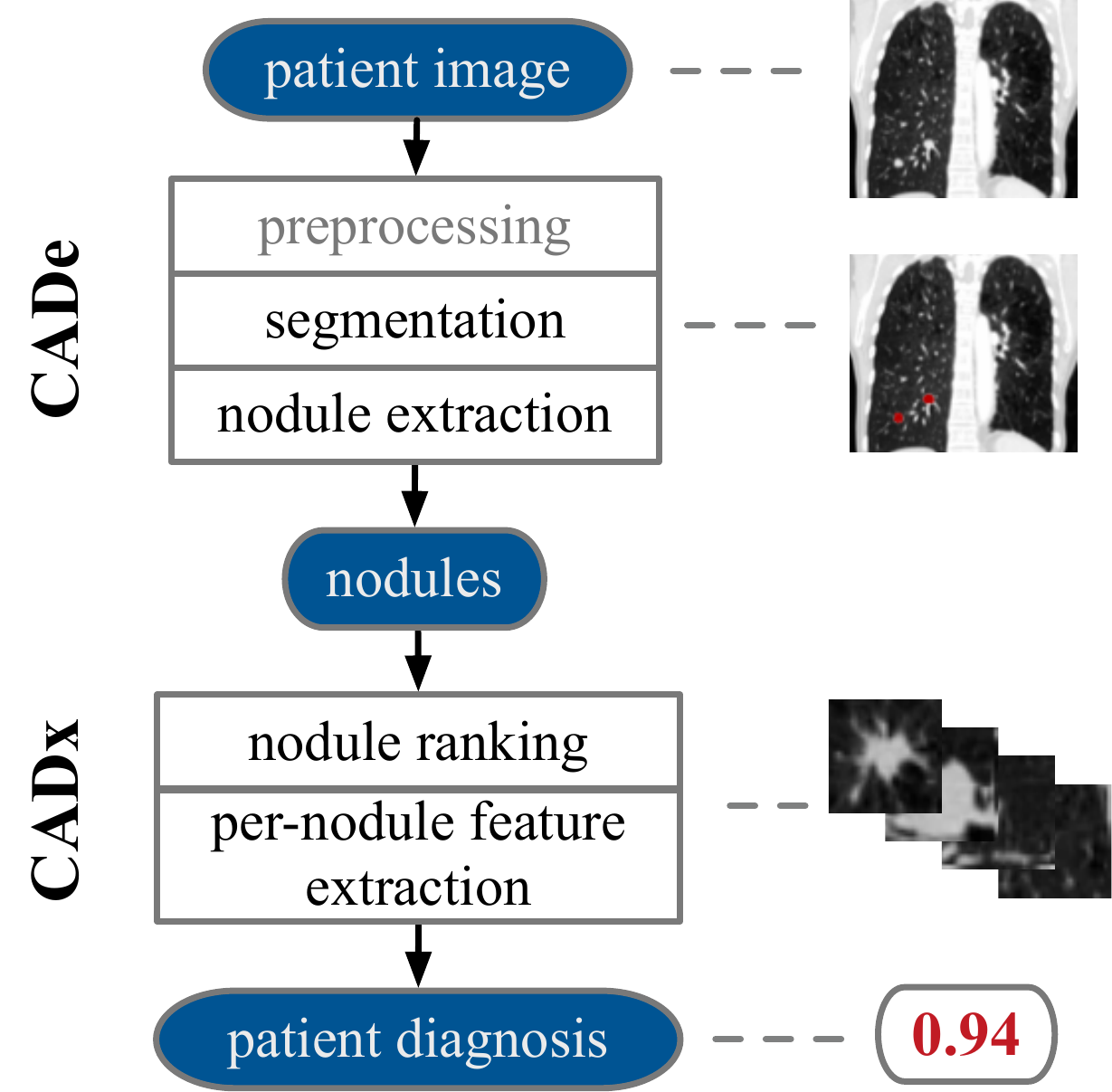}
\end{center}
\caption{Our overall CAD system diagram. Since CADx performance is so reliant on the quality of the nodule candidates generated by the CADe, both were developed simultaneously to achieve the best overall performance.}
\label{fig:sys_diag}
\end{figure}

Here we introduce our CADe/CADx system and provide an overview of the key components. The details pertinent to each individual component are provided in Sections \ref{sec:cade} and \ref{sec:cadx}. The overall system diagram is shown in Figure \ref{fig:sys_diag}. Our system takes a raw 3D CT scan of the lung as input and provides as outputs a per-patient malignancy classification probability, per-nodule malignancy scores, and segmented lung nodule candidates. The system can refer patients or nodules whose results have a high degree of uncertainty to a radiologist for confirmation.

As a first step, we preprocess each full 3D scan to a consistent format, described in further detail in Section \ref{sec:data}. The preprocessed 3D scan is then fed into the CADe module, which performs 3D segmentation. The goal of our CADe module is to identify and localize lung nodules with the highest possible recall. The output of CADe module is a list of identified lung nodules which are then fed into our CADx module. CADx module uses two cascaded 3D deep learning models. The first model ranks the candidates based on their malignancy risk. The second model then uses that ranking to select the top-$k$ candidates and perform multiple-instance classification to make a patient diagnosis. 

The full system is developed and tuned simultaneously, since CADx performance is dependent on CADe performance. This is different than current state-of-the-practice where CADe and CADx components are treated as independent components and optimized separately. As we show quantitatively in Section \ref{sec:results}, CADe approaches optimized for performance on the LUNA16 benchmark may not perform as well when used as part of a full CADe/CADx system. In particular, when the CADx system is trained with candidates with a high FPR, it becomes much more robust to false positive candidates, resulting in improved performance regardless of the underlying CADe system's FPR (see Section \ref{sec:cadecadxres}.)

\section{Data}
\label{sec:data}

\subsection{Lung CT Datasets}
Our lung cancer system is based on two publicly-available low-dose CT scan datasets. The first is the LIDC-IDRI \cite{armato2011} dataset, which contains lesion annotations from four experienced thoracic radiologists. This dataset is used for both our CADe and CADx components. We use the curated version of this dataset provided as part of the LUNA16 challenge \cite{luna16}, which includes 888 patients and consensus labels based on the agreement of 3/4 radiologist annotators, resulting in a total of 1186 annotated nodules. We further associate the malignancy scores given by each of the annotating radiologists to these consensus nodule labels. Unfortunately, the LIDC-IDRI annotations do not include very large nodules or masses, which are important for full patient diagnosis. This is an important limitation of this dataset for use in a standalone CADe system.

The second dataset we use is the one provided by the National Cancer Institute for the 2017 Data Science Bowl on Kaggle \cite{dsb}. It has two separate sets of CT scans, namely Stage-1 and Stage-2, which were used as training and test sets for the Kaggle challenge. Stage-1 data consists of 1595 patients, out of whom 419 (${\sim}26\%$) were diagnosed with lung cancer within one year after the CT scans were acquired. Stage-2 includes 506 patients, out of whom 153 (${\sim}30\%$) were diagnosed with lung cancer. Stage-2 data is generally more recent and higher quality (thinner slice thickness) which helps test the ability of CADx systems to generalize beyond their training data. Kaggle data was used for the training (Stage-1) and testing (Stage-2) of the CADx component of our system.

\subsection{Data Preprocessing and Augmentation}
Here we provide details on the data preprocessing and augmentation steps performed prior to CADe/CADx model training.
CT scan preprocessing involved only clipping the scan image range to between -1000 and 400 Hounsfield units in order to remove most of the variation due to bone and resampling the images such that voxels measured 1 mm in each dimension. Images were normalized to have a mean voxel value of 0 and variance of 1 before being input to the neural networks, as is standard.

\begin{figure}
\begin{center}
\includegraphics[width=0.48\textwidth]{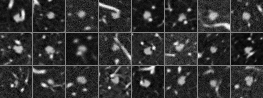}
\end{center}
\caption{Randomly sampled augmentations of a single nodule demonstrating our extensive augmentation transforms used during model training.}
\label{fig:nodule_aug}
\end{figure}

We used extensive data augmentation (shown in Figure \ref{fig:nodule_aug}) to reduce overfitting and maximize the transfer learning across images taken with different setups. Data augmentation was used during training of all of our models, but not used at test time. Our affine transform augmentation was made up of uniformly sampled 3D rotations and reflections, as well as smaller random scaling from $\mathcal{N}(0,0.06)$\% and translations from $\mathcal{N}(0, 1)$ mm, independently in all three dimensions. Additional image transforms included random gamma transformations sampled from $\mathcal{U}(0.7, 1.3)$, Gaussian blur or unsharp masking with $\sigma$ sampled from $\mathcal{U}(0, 1.5)$, and additive Gaussian noise with $\sigma$ sampled from $\mathcal{N}(0, 0.03)$. To ensure the highest possible CADe recall for large nodules, additional aggressive scaling augmentation ($3\times$ scaling distribution above, for upsampling) was used in the training of that component. This allowed our CADe system to find nodule candidates significantly larger than any actually annotated in the LIDC-IDRI dataset. 

\section{CADe}
\label{sec:cade}

Our CADe system was developed to have as high of a sensitivity/recall as possible for identifying and localizing pulmonary nodules while still keeping the number of false positives low enough to be manageable for our CADx system. It is able to analyze any size CT scan and uses only local visual information to make its decisions. Our CADe system is made up of 1) a 3D segmentation network, which labels every voxel of the CT scan with a nodule probability, and 2) a 3D scoring network, which computes refined nodule probability estimates for full nodule candidates generated from the segmentation and allows for both greater interpretability and false positive reduction. The following subsections explain these networks in detail.

\subsection{Segmentation for Candidate Extraction}

Our primary nodule segmentation network is a 3D fully-convolutional neural network based on the V-Net architecture, which has been demonstrated to be effective for 3D medical image segmentation \cite{vnet_3dv16}. Our architecture\footnote{We began with the V-Net architecture as described in the paper \cite{vnet_3dv16} and tuned the number of layers, number of features, non-linearity functions, and normalization layers based on performance on the LUNA16 dataset.} uses three encoder-decoder block pairs, with corresponding skip connections, in addition to the input and output blocks. Our encoder blocks are made up of a $2\times$ downsampling convolution, two layers of kernel size 3 convolutions, and residual connection to the output. Decoder blocks are the same, but with a $2\times$ upsampling deconvolution. The innermost encoder-decoder block pair include channel-wise dropout between the sampling convolution and the two main convolutional layers. All blocks use instance normalization \cite{ulyanov16} instead of batch normalization as well as ReLU nonlinearities.

For training, we use the LUNA16 pulmonary nodule annotations, which provide nodule centers and diameters, to annotate individual voxels of CT scans with spherical masks. Since the number of voxels corresponding to the nodule annotations is very small compared to the total number of voxels, and each CT scan is too large to realistically use to train our network, we sample $64^3$-shaped blocks from the original images. For each patient, we sample blocks near known nodules with a probability of $1-0.7^N$ and sample randomly from the image the with a probability of $0.7^N$, where $N$ is the number of nodules for that patient. We employed this random sampling strategy in order to ensure that the network is exposed to true nodules a sufficient number of times as well as a diverse selection of background. We train the network with a cross entropy loss function that weights voxels within a nodule twice as much as background voxels. Each patient scan is seen once per epoch, regardless of how many nodules are contained within. Our network is trained with 16 block samples per batch using the Adam optimizer \cite{adam_iclr15} with learning rate of $10^{-3}$ for 2500 epochs.

As full preprocessed CT scan images are too large to fit into GPU memory\footnote{We used an NVIDIA Quadro GP100 or equivalent for all model training and evaluation described in this paper.} while being passed through the trained nodule segmentation model at test time, we split images into eight $256^3$-shaped overlapping blocks and stitch the output segmentations together, weighting voxels in the overlapping edge regions by how much of their field-of-view was contained within the borders of their respective input blocks. This results in a full image segmentation that is as good as if the entire image were evaluated by the network in one pass on a CPU, but requires an order of magnitude less time.

\begin{figure}
\begin{center}
\includegraphics[width=0.48\textwidth]{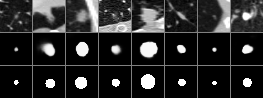}
\end{center}
\caption{Sample nodule segmentations from our CADe segmentation model, sliced through the center of each nodule candidate. First row: Input CT scan images from LIDC-IDRI test data. Second row: Our corresponding segmentation probabilities. Third row: (Spherical) voxelwise labels extracted from the LUNA16 annotations.}
\label{fig:seg_examples}
\end{figure}

CADe candidates are generated from the full segmented image by thresholding the output voxel scores, applying a nearest-neighbors binary opening filter, and labeling all separate regions based on a voxel connectivity of one. The center of each nodule candidate is taken to be the center of mass of the voxel scores of each constituent voxel after thresholding. Sample segmentations of generated candidates on LUNA16 test data are shown in Figure \ref{fig:seg_examples} alongside the original spherical annotation.

\subsection{False Positive Reduction for LUNA16 Evaluation}

All generated candidates are supplied to the CADx system. However, for the purposes of evaluating our CADe system on the LUNA16 benchmark, we developed a scoring network that operates on $32^3$-shaped blocks around the candidate center. The architecture of this network is shown in Figure \ref{fig:cnn_arch}. The network was trained with the SGD optimizer (learning rate 0.01, momentum 0.9, and $0.8\times$ learning rate decay every 100 epochs) with batches of 16 candidates for 2500 epochs, with true nodule candidates weighted twice as heavily as false positive candidates in the cross-entropy loss. As we show in Section \ref{sec:results}, our scoring network can be used to dramatically reduce the number of false positives found by the segmentation network.

Our candidate scoring network outputs include model uncertainty as quantified by Monte Carlo (MC) dropout \cite{gal_iclr16, gal_icml16}. We show in Section \ref{sec:referral} that the addition of this model uncertainty makes these scores well-calibrated to the LUNA16 data distribution and thus the probabilities output by the network for each candidate are directly interpretable as real probabilities of being a nodule.

\section{CADx}
\label{sec:cadx}

Our CADx system classifies each CT scan as malignant or benign while associating malignancy scores to each individual candidate (i.e., suspicious lung nodule) that it processes. The CADx system is composed of two consecutive 3D convolutional neural networks. The first network is a regression network that we use to rank candidates for each CT scan based on their malignancy risks. As each CT scan may have a large number of candidates, the ranking step enables us to reduce noise and focus on candidates that are more likely to be malignant. The second network performs classification by processing the top-$k$ nodule candidates ranked by the first network using a multiple instance learning approach. We refer to the first and the second networks as \emph{malignancy ranking} and \emph{malignancy classification} networks, respectively. 

\subsection{Malignancy Ranking Network}

Our malignancy ranking network architecture has the same basic structure as the CADe candidate scoring network, described in Figure \ref{fig:cnn_arch}. To train our ranking network, we regress a label derived from the malignancy scores from the LIDC-IDRI dataset \cite{armato2011} provided by four different radiologists. In these annotations, each radiologist assigned an integer malignancy score between $1$ and $5$ for each significant nodule they were able to identify ($1$ and $5$ indicating the least and the most malignant scores, respectively). As a result, each nodule has between one and four scores. These scores are subjective in that they were not confirmed by follow-up biopsies nor were they consistent between different radiologists. Therefore, they merely provide noisy labels which we use to train our malignancy ranking network for the purpose of ranking candidates detected in each CT scan. To reduce noise in the training data, we selected nodules that were scored by at least three radiologists and averaged their scores to come up with final nodule malignancy scores for training. We assigned a score of $1$ for candidates that were either scored by fewer than three radiologists, were too small to be scored at all, or were false positive detections.

Eight subsets of the LUNA16 dataset were selected for training the malignancy ranking and two subsets were selected for development. During initial training, we randomly select nodules with at least three radiologist scores for the first 50 epochs. After that, we randomly select those nodules 90\% of the time and sample randomly from the remaining set, which only has nodules of score $1$, 10\% of the time. This can be seen as a curriculum learning strategy to prevent the network from biasing towards benign nodule scores since most of the detected candidates are benign. We train the network using the Adam optimizer \cite{adam_iclr15} with a learning rate of $10^{-4}$ with mean absolute error as the loss function for 750 epochs with a batch size of $32$ candidates. The ranking network training and hyperparameter tuning was performed using a library of LUNA16 nodule candidates created by our earlier work on standalone CADe \cite{ozdemir2017propagating}.

The malignancy ranking network generates a list of ranked candidates for each CT scan based on their malignancy risks. These ranked candidates are then used to train or be evaluated by the malignancy classification network.

\begin{figure}
\begin{center}
\textbf{Base Neural Network Model}\\[.2cm]
\small
\begin{tabular}{ c | c | c  }
	Type & Filters & Output \\ \hline
	Conv. Block & 32 &  16$\times$16$\times$16\\
	Conv. Block & 64 & 8$\times$8$\times$8 \\
	Conv. Block & 128 & 4$\times$4$\times$4 \\
	Fully Connected + PReLU&  & 1024 \\
	Dropout &  & 1024 \\
	Fully Connected  &  & 1 \\
\end{tabular}
\vspace{.4cm}\\
\normalsize{\textbf{Convolutional Block}}\\[.2cm]
\small
\begin{tabular}{ c | c   }
        Type & Size/Stride \\ \hline
        Convolutional + PReLU& 5$\times$5$\times$5/1 \\
        Batch Normalization &  \\
        Maxpool & 2$\times$2$\times$2/2  \\
\end{tabular}
\end{center}
\caption{Base neural network architecture used for nodule candidate scoring, malignancy ranking, and multiple-instance malignancy classification. The architecture hyperparameters were found through experimentation on the two CADx tasks, namely malignancy ranking and classification.}
\label{fig:cnn_arch}
\end{figure}

\subsection{Malignancy Classification Network} \label{sec:malig-class}

Patient-wide malignancy classification is the final step in our CADe/CADx pipeline. Our malignancy classification network processes a predefined number of ranked nodule candidates and outputs a probabilistic malignancy score for each CT scan. It also assigns a non-probabilistic malignancy risk score for each candidate that it processes. 

We use an attention-based multiple instance learning (MIL) framework to train our malignancy classification network \cite{ilse_icml18}. The MIL framework is based on a convolutional neural network shared by all selected candidates followed by a combination layer that combines features of each candidate using an attention mechanism. As our shared network, we use the same basic architecture described in Figure \ref{fig:cnn_arch} with the last fully connected layer removed. The shared network generates a feature vector $\mathbf{h}_i \in \mathbb{R}^{1024\times1}$ for each candidate $i$. These feature vectors are then combined via the following attention model:
\beq
\mathbf{x}_i = \tanh (\mathbf{W}_1 \mathbf{h}_i) 
\eeq
\beq\label{eq:attn-weights}
\mathbf{y} = \textrm{softmax} (\mathbf{w}_2^T \mathbf{X}) 
\eeq
\beq
p = \textrm{sigmoid} (\mathbf{w}_3^T \mathbf{H} \mathbf{y}) \label{eq:attn_sigmoid}
\eeq
where $\mathbf{W}_1 \in \mathbb{R}^{128\times1024}$, $\mathbf{w}_2 \in \mathbb{R}^{128\times1}$, and $\mathbf{w}_3 \in \mathbb{R}^{1024\times1}$ denote the learned weights of the attention-based combination model; and $\mathbf{X} = [\mathbf{x}_1, \ldots, \mathbf{x}_k]$ and $\mathbf{H} = [\mathbf{h}_1, \ldots, \mathbf{h}_k]$ denote the concatenated feature vectors from the $k$ candidates. Note that we have omitted the bias terms above to keep the notation simple, though they exist in the actual implementation.

The attention mechanism allows the model to learn permutation-invariant weights as a function of the feature vectors, denoted by $\mathbf{y}$ in \eqref{eq:attn-weights}, where $y_i \in [0,1]$ and $\sum_i y_i = 1$. These weights are then used to compute a weighted average of the feature vectors themselves, which is fed into a fully connected layer followed by a sigmoid nonlinearity to compute a probabilistic malignancy score for each patient, as shown in \eqref{eq:attn_sigmoid}. We explored other combination methods such as Noisy-OR (NOR) \cite{viola_nips05}, Leaky NOR (L-NOR) \cite{kaggle17}, and log-sum exponentiation (LSE) \cite{ramon_icml00}. We found that they did not work as well as the attention-based model for our dataset and they were harder to optimize as L-NOR and LSE introduced additional hyperparameters.    

For training the malignancy classification network, we combined the Kaggle Stage-1 dataset with a subset of the LUNA16 dataset as our training and development data, and set aside Kaggle Stage-2 dataset as our test data. Specifically, we selected the LUNA16 patients having a nodule with at least three radiologist scores and an average score $\geq4$ as positive (malignant), and patients having no radiologist annotations or patients with all their nodules having an average score $\leq2$ as negative (benign). This procedure added 556 patients to our training data, out of whom 169 (${\sim}30\%$) were labeled malignant. As a result, we ended up with a total of $2101$ patients ($556$ from LUNA16 and $1595$ from Kaggle Stage-1) for training. 

One limitation of the LUNA16 dataset is that the radiologist annotators were specifically instructed to ignore nodules that were larger than $30$ mm. As a result, machine learning based CADe systems trained with the LUNA16 dataset could learn to ignore large nodules even when aggressive data augmentation (specifically zooming) techniques are employed during training. Nodule size, however, is an important indicator for malignancy, with larger nodules having a much higher likelihood of being malignant. To alleviate this problem, we use two sets of candidates from the CADe system. The first set has candidates from the original isotropically sampled ($1\text{ mm})^3$/voxel scans. To be able to detect large nodules, we created a second candidate set by downsampling each scan to $(2\text{ mm})^3$/voxel and passing them through our trained CADe system. These candidates are then similarly passed through our trained malignancy ranking network to create a ranked list of downsampled candidates.

To train the malignancy classification network, we select the top-$k$ candidates from each ranked candidate list (the original and downsampled) where we treat $k$ as a hyperparameter to optimize. For patients with less than $k$ candidates, we use the existing number of candidates, which does not affect our training procedure. We set aside $300$ patients for development and used the remaining set for training. Our network was trained using the SGD optimizer with momentum $0.9$ and an initial learning rate of $0.01$,  which was decreased by a factor of $2$ every 50 epochs. We used binary cross entropy as the loss function and trained the network with a batch size of $32$ scans for 750 epochs. Our best performing CADx model (as shown by Figure \ref{fig:cadx_roc}) uses a total of $4$ candidates, i.e., top-$2$ from each of the original and downsampled candidate lists.

We obtained model uncertainty estimates using a combination of the MC dropout \cite{gal_icml16} and the deep ensembles method \cite{blundell_nips17} by training five different models with different train and development dataset splits. We show in the next section that this method provides calibrated malignancy probabilities which can be interpreted as true probabilities and  used as a reliable risk-utility metric for subsequent decision making in clinical settings.

\begin{figure}
	\begin{center}
		\includegraphics[width=0.48\textwidth]{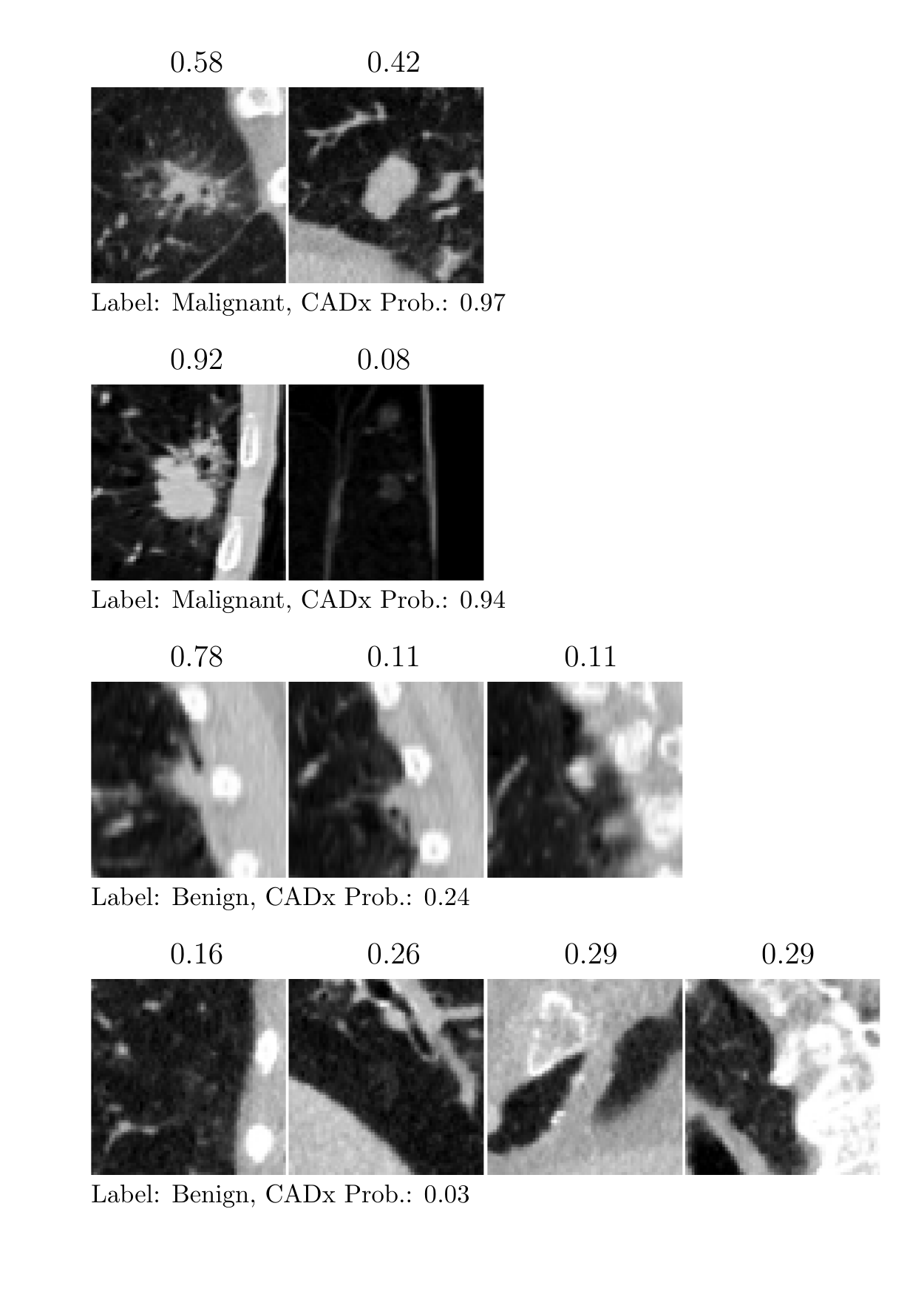}\vspace{-1cm}
	\end{center}
	\caption{Example nodule candidates with CADx malignancy probabilities along with corresponding candidate attention weights. Each row represents candidates from a specific patient. The scores on top of each candidate are the corresponding CADx network attention weights, which sum up to $1$ and represent how much each candidate contributes relatively to the final CADx score. Estimated patient-level CADx malignancy probability is given in the bottom of each row.}
	\label{fig:cadx_examples}
\end{figure}

For visualization, we show nodule candidates from a few example patients in Figure \ref{fig:cadx_examples}. The corresponding attention weights shown on top of each row represent how much each candidate contributes to the overall malignancy prediction probability for a given patient. One can interpret these weights as relative malignancy scores for patients with high estimated malignancy probabilities.

\section{Results}
\label{sec:results}

We evaluate our CADe system on the LUNA16 benchmark and our CADx system on the Kaggle Stage-2 test set. Since CADx directly relies on CADe, the success of the CADx system acts as additional validation of the CADe system and its ability to generalize to an independent dataset. Likewise, the CADx system is trained and validated on the Kaggle Stage-1 dataset but tested on the Kaggle Stage-2 dataset, which is more recent and has different image quality.

\subsection{CADe Results}

To evaluate our CADe system on the LUNA16 benchmark, we use 10-fold cross validation with the prescribed splits and provided lung volume masks. For each test split, another split was used as a validation set to select both the corresponding segmentation model and scoring model checkpoints, while the other eight were used to train both models. Candidate selection from the segmented test scans using a segmentation threshold of $0.2$ achieved a sensitivity (or recall) of 96.5\%  with 19.7 average false positives per scan without any false positive reduction step. This false positive average is dominated by candidates from a small number of patients with large regions of uncertain segmentation. 

\begin{figure}[h]
	\begin{center}
		\includegraphics[width=0.48\textwidth]{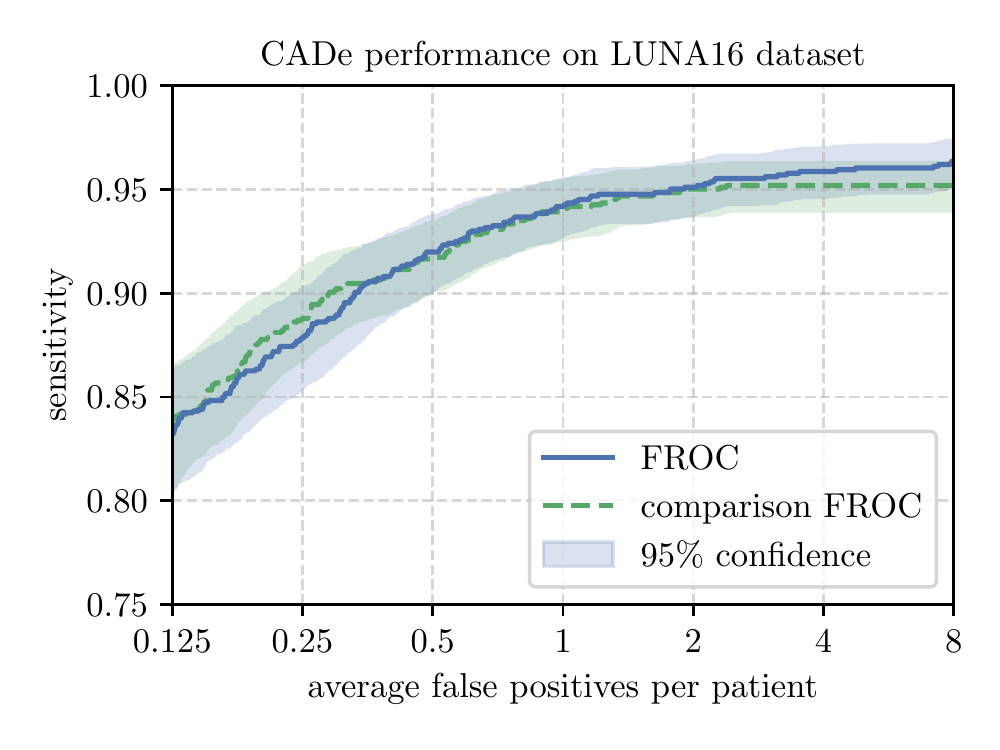}\vspace{-.5cm}
	\end{center}
        \caption{The free response operating characteristic (FROC) for our CADe candidate generation and scoring system on the LUNA16 dataset, with patient-bootstrapped 95\% confidence interval, and the same results with our comparison model architecture (3D U-Net.)}
	\label{fig:cade_froc}
\end{figure}

\begin{figure}[h]
	\begin{center}
		\includegraphics[width=0.48\textwidth]{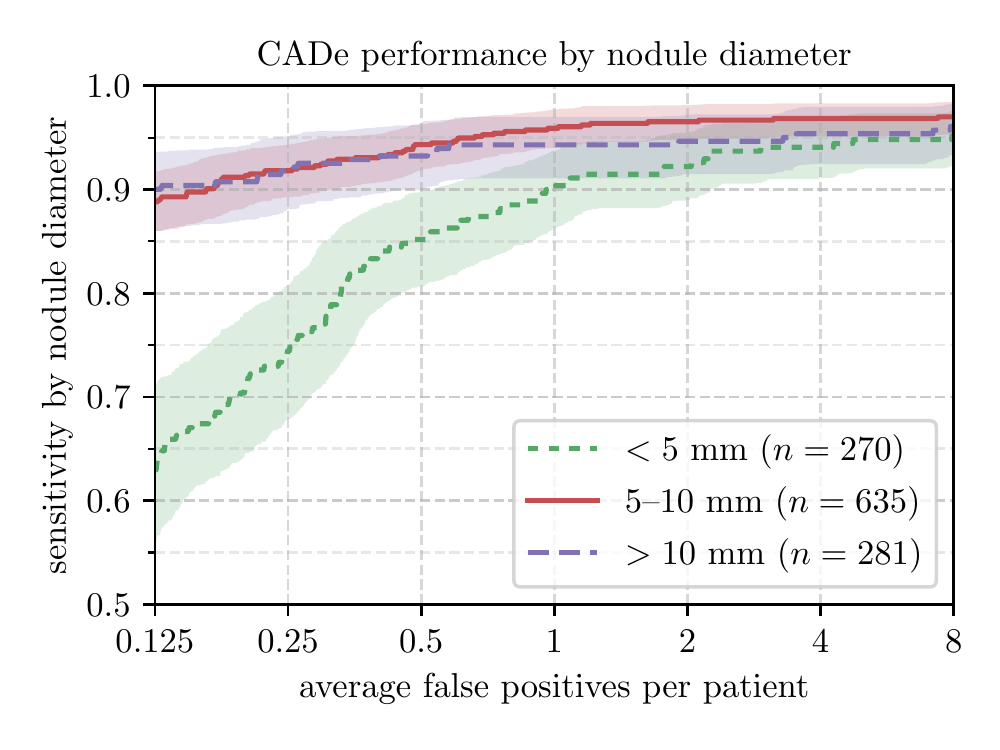}\vspace{-.5cm}
	\end{center}
	\caption{CADe FROC on LUNA16 data, breaking down the sensitivity by nodule diameter. CADe sensitivity for the smallest group of nodules (between 3 mm and 5 mm diameter) is significantly worse than the sensitivity for larger nodules (between 5 mm and 30 mm diameter) at lower thresholds that correspond to reduced false positives.}
	\label{fig:cade_froc_diam}
\end{figure}

Our CADe scoring network allows a dramatic reduction in the number of false positives, while keeping the nodule sensitivity extremely high. Our CADe sensitivity as a function of false positive rate per scan is shown in Figure \ref{fig:cade_froc} and the results on the LUNA16 metric are shown in Table \ref{tab:main_comparison}. Our average LUNA16 metric of 0.921 is comparable to or better than state-of-the-art published results \cite{bagci_miccai18}, even though our CADe is not optimized for the high-precision limit. The breakdown by nodule diameter in Figure \ref{fig:cade_froc_diam} shows that our CADe performance is strongest on pulmonary nodules with a diameter greater than 5 mm. This is primarily because our false positives are dominated by small candidates which are difficult to distinguish from true positive small nodules.

In addition to the results using our primary V-Net based segmentation architecture, we generate comparison results from our full training and evaluation pipeline using the standard 3D U-Net architecture \cite{cciccek20163d}, with the number of features per block tuned for validation performance on the LUNA16 dataset. As seen in Figure \ref{fig:cade_froc} and Table \ref{tab:main_comparison}, the CADe and CADx results using this comparison architecture are similar to our primary results, indicating that the success of our CADe system is based more on the sampling and data-augmentation techniques used in the training pipeline than the architectural details of the neural network.


\begin{table*}[tb]
\centering
\caption{\label{tab:main_comparison}CADe and CADx results by CADe threshold false positive rate}
\vspace{-.2cm}
\begin{tabular}{ c | c | c || c | c }
    & \multicolumn{2}{c ||}{LUNA16 CADe sensitivity} & \multicolumn{2}{c}{Kaggle Stage-2 CADx AUROC} \\ \hline
    CADe & Primary arch. & Comparison arch. & \multicolumn{2}{c}{Evaluated on nodule cands.\ from} \\
    FP / patient & (V-Net based) & (3D U-Net) & Primary arch. & Comparison arch. \\ \hline
       $1/8$ & 0.832 & 0.839 & 0.867 & 0.869 \\
       $1/4$ & 0.879 & 0.888 & 0.868 & 0.868 \\
       $1/2$ & 0.920 & 0.917 & 0.863 & 0.863 \\
       1 & 0.942 & 0.941 & 0.868 & 0.863 \\
       2 & 0.951 & 0.950 & 0.866 & 0.858 \\
       4 & 0.959 & 0.952 & 0.869 & 0.858\\
       8 & 0.964 & 0.952 & \textbf{0.869} & 0.859 \\ \cline{1-3}
       average & \textbf{0.921} & 0.920 \\
\end{tabular}
\end{table*}

\subsection{CADx Results}

\begin{figure}
        \begin{center}
                \includegraphics[width=0.48\textwidth]{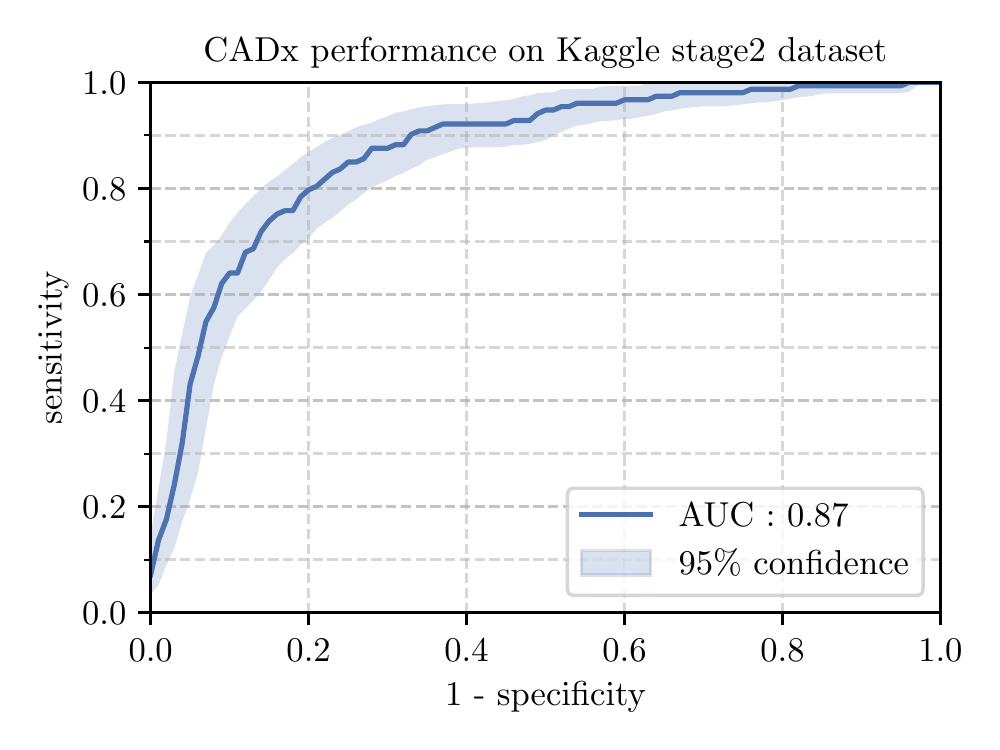}\vspace{-.5cm}
        \end{center}
        \caption{The receiver operating characteristic (ROC) curve for our CADx system on the Kaggle Stage-2 test set, trained on both LUNA16 and Kaggle Stage-1 data, with patient-bootstrapped 95\% confidence interval.}
        \label{fig:cadx_roc}
\end{figure}

Our best performing CADx model was trained and evaluated with nodule candidates obtained by setting the CADe segmentation threshold to $0.5$, which resulted in approximately $8$ false positives per scan on average without any false positive reduction step. This kept spurious candidates to a reasonable level while not removing large nodule candidates not annotated in LUNA16. This corresponds to a CADe segmentation $F_1$ score (Dice coefficient) for LUNA16 of 0.40, with 0.25 precision and 0.93 recall.

We trained our CADx model with the top-$4$ ranked candidates per patient as it provided the best performance on our development set. We evaluated our final CADx system on the Kaggle Stage-2 data, which was set aside as our test set. As shown in Figure \ref{fig:cadx_roc}, our CADx system achieved an average ROC Area Under the Curve (AUC) of $0.87$ on this dataset. This performance is the same as the winning Kaggle solution \cite{kaggle17}, but unlike that solution, did not require us to hand-label the nodules in the Kaggle Stage-1 data nor ensemble different CADx solutions as was done by other top-10 Kaggle solutions.

\subsection{CADe-CADx Interaction Studies}
\label{sec:cadecadxres}

\begin{table}[tb]
\centering
\caption{\label{tab:train_test_tradeoff}CADx results by CADe threshold for training and testing}
\vspace{-.2cm}
\begin{tabular}{ r c | c c c }
    & FP / & \multicolumn{3}{c}{Trained on} \\
    & patient & $1/8$ & 1 & 8 \\ \cline{2-5} \rule{0pt}{.35cm}
    \multirow{3}{*}{\rotatebox[origin=c]{90}{Tested on}}\hspace{-.15cm}
    & $1/8$ & 0.847 & 0.848 & 0.867 \\[.1cm]
    & 1 & 0.815 & 0.852 & 0.868 \\[.1cm]
    & 8 & 0.753 & 0.833 & \textbf{0.869} \\[.1cm]
\end{tabular}
\end{table}

We further studied the effect of CADe false positive reduction by evaluating our CADx system with Kaggle Stage-2 candidates filtered by our CADe false positive reduction model at the LUNA16 evaluation points, shown in Table \ref{tab:main_comparison}. We again found that our CADx system is relatively insensitive to false positive candidates from the CADe system and that false positive reduction steps add no value.

Additionally, we explored the CADe-CADx coupling in candidate generation in both training and testing stages, shown in Table \ref{tab:train_test_tradeoff}. In addition to our original CADx results trained using Kaggle Stage-1 candidates with a CADe threshold equivalent of $8$ FP per scan, we fully retrained our CADx system using candidates filtered at the $1$ and $1/8$ FP per scan LUNA16 evaluation points. We found that, not only did the CADx AUROC degrade significantly, but these CADx systems were much less robust to false positives. These results show that the presence of false positive candidates actually benefits CADx training by improving its internal CADe capabilities, and thus robustness to false positives.

\section{Calibration \& Referral}
\label{sec:referral}

In order to generate meaningful probability estimates from both our CADe and CADx components, we include Bayesian-motivated model uncertainty and verify that the consequent estimates are well calibrated on test data. Well-calibrated probability estimates are useful because they give us the ability to interpret the nodule candidate and overall malignancy scores as true probabilities and use them for subsequent decision making. 

Model uncertainty captures the effect of uncertainty in the neural network parameters (weights and biases) and shrinks as the model is trained on more, representative data. Model uncertainty is particularly significant in deep learning, where models often have millions of free parameters. By quantifying and including model uncertainty, our system can more reliably assess out-of-cohort data without making overly-confident predictions, making it more trustworthy for real-world usage.

\begin{figure}
\begin{center}
\includegraphics[width=0.48\textwidth]{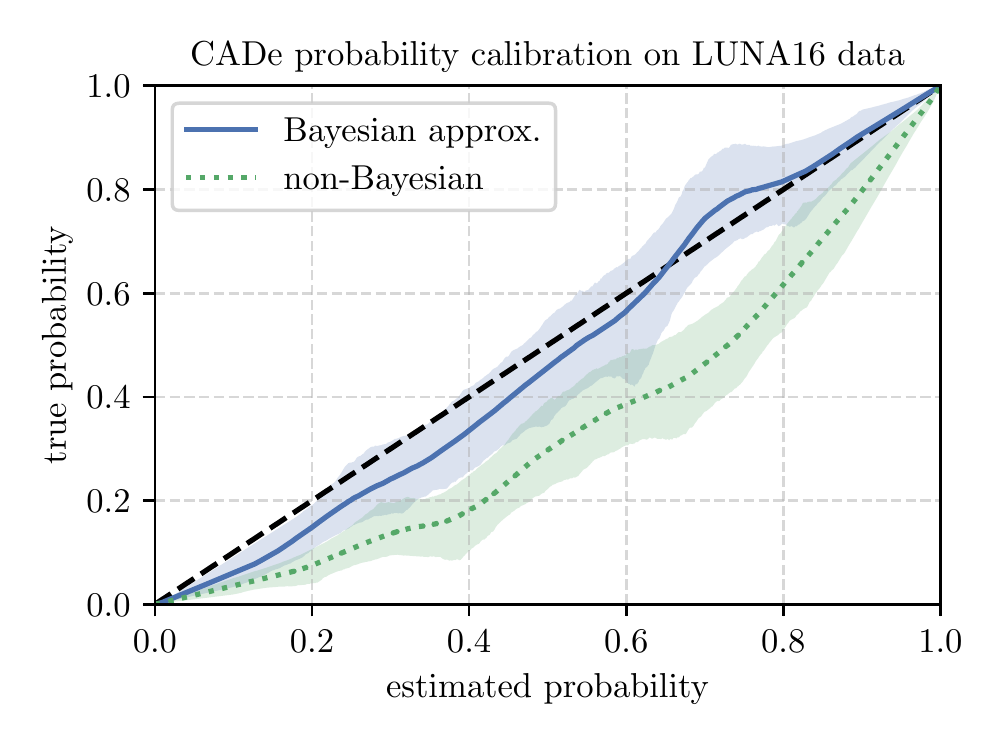}
\end{center}
    \caption{Probability calibration curves for the Bayesian approximation and non-Bayesian (standard) variants of the CADe candidate scoring neural networks, with patient-bootstrapped 95\% confidence intervals. The estimated nodule probabilities output by the CADe scoring network are well-calibrated when model uncertainty is included.}
\label{fig:cade_calib}
\end{figure}

For the nodule candidate scoring network, we trained 10 models using each of the 10 LUNA16 data subsets as a test set. Model uncertainty was approximated by using Monte Carlo (MC) dropout \cite{gal_iclr16, gal_icml16}. This technique randomly drops out features during both training and testing, allowing us to perform approximate Bayesian inference by sampling over these features. The probabilities produced by the candidate scoring networks thus incorporate uncertainty in the model weights in addition to uncertainty inherent to the data including label noise. In Figure \ref{fig:cade_calib}, this is shown to produce well-calibrated probability estimates on the LUNA16 test data. When model uncertainty is not incorporated, by generating estimates without dropout at test time as is typical, the probability estimates from the network tend to be overconfident in their predictions, as can be seen in the figure. We expect this difference would be even more dramatic for evaluation on data with significant domain shift (for example, due to unusual hardware, procedures or patients). 

\begin{figure}
\begin{center}
\includegraphics[width=0.48\textwidth]{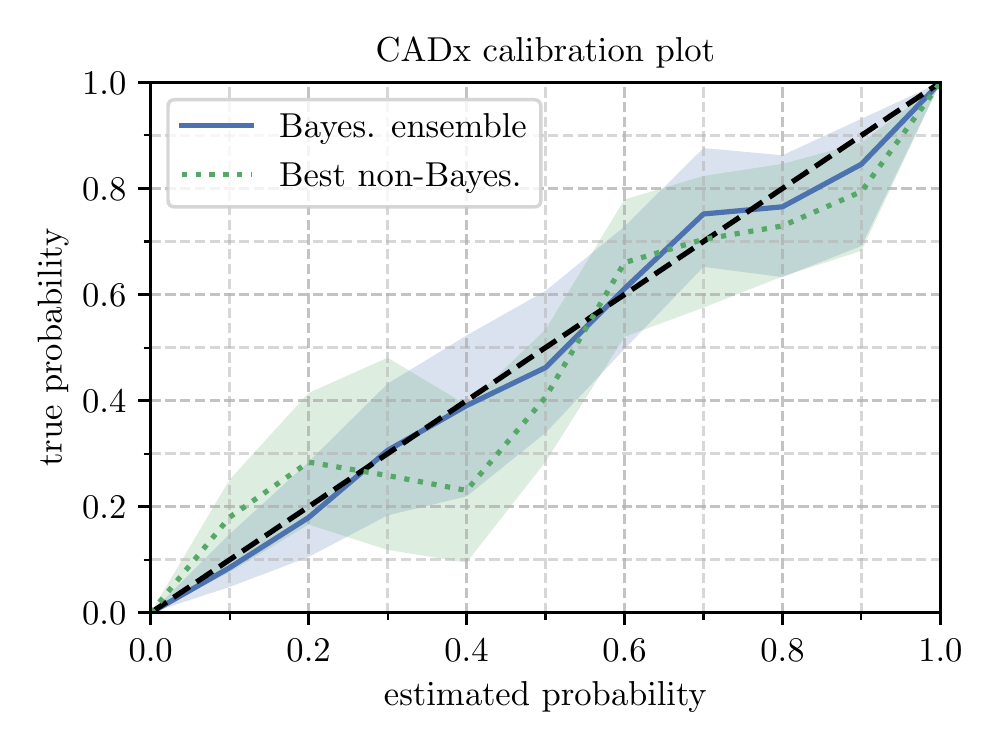}\vspace{-.5cm}
\end{center}
\caption{Probability calibration curves for the Bayesian ensemble and the best non-Bayesian CADx model, with patient-bootstrapped 95\% confidence intervals. The estimated malignancy probabilities output by the CADx Bayesian ensemble are well-calibrated when model uncertainty is included.}
\label{fig:cadx_calib}
\end{figure}

As mentioned in Section \ref{sec:malig-class}, we used a combination of the MC dropout \cite{gal_icml16} and deep ensembles \cite{blundell_nips17} methods to assess model uncertainty of our malignancy classification network. The model ensembling method simply involves training a model multiple times, starting with randomly initialized parameters, and averaging their results. We trained the malignancy classification network on five different train and development splits and used the predictions, obtained by MC dropout, from these models to assess model uncertainty as well as boost the performance. Since we only had one test set, it was more effective to estimate model uncertainty using the ensemble method combined with MC dropout, which has been shown in some cases to produce more reliable uncertainties than the MC dropout alone \cite{blundell_nips17}. The calibration curves are shown in Figure \ref{fig:cadx_calib}. While the best individual network did not systematically overestimate uncertainties like was seen for the CADe candidate scoring network, the ensemble probabilities appear to be more stably calibrated over the full range of probabilities.

\begin{figure}
\begin{center}
\includegraphics[width=0.48\textwidth]{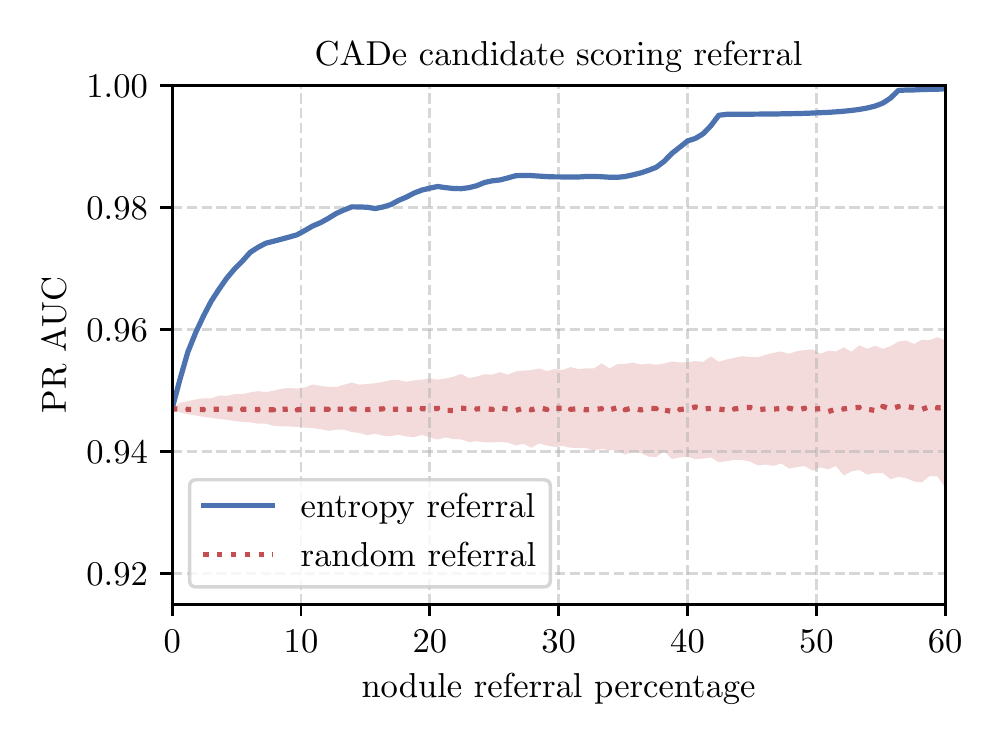}\vspace{-.5cm}
\end{center}
\caption{The CADe scoring area under the precision-recall curve as a function of referral percentage, for the entropy referral strategy and a random strategy (with 95\% confidence interval).}
\label{fig:cade_ref}
\end{figure}

We explored the potential of using these well-calibrated probabilities to improve the reliability of our CAD systems through referral strategies, in which a radiologist could check the most critical subset of patients and nodules. We found that referral based on highest entropy,
\beq
H(p) = -p\ln(p)-(1-p)\ln(1-p),
\eeq
was advantageous, particularly for CADe nodule referral. In Figure \ref{fig:cade_ref}, the area under the precision-recall curve for our nodule candidate CADe scoring is plotted as a function of the percentage of nodules ignored based on this referral strategy. There is a dramatic improvement in performance, even at a referral rate below 10\%, and the performance on remaining nodules monotonically approaches 1 as the referral rate is increased. This further demonstrates that the nodule candidate scoring probabilities are well calibrated even extremely close to 0 and 1. The nodule candidate scoring system combined with a ``perfect'' radiologist able to evaluate a subset of candidates could nearly completely eliminate false positive nodules from the CADe candidate generation process without missing additional true positive nodules.

\begin{figure}
\begin{center}
\includegraphics[width=0.48\textwidth]{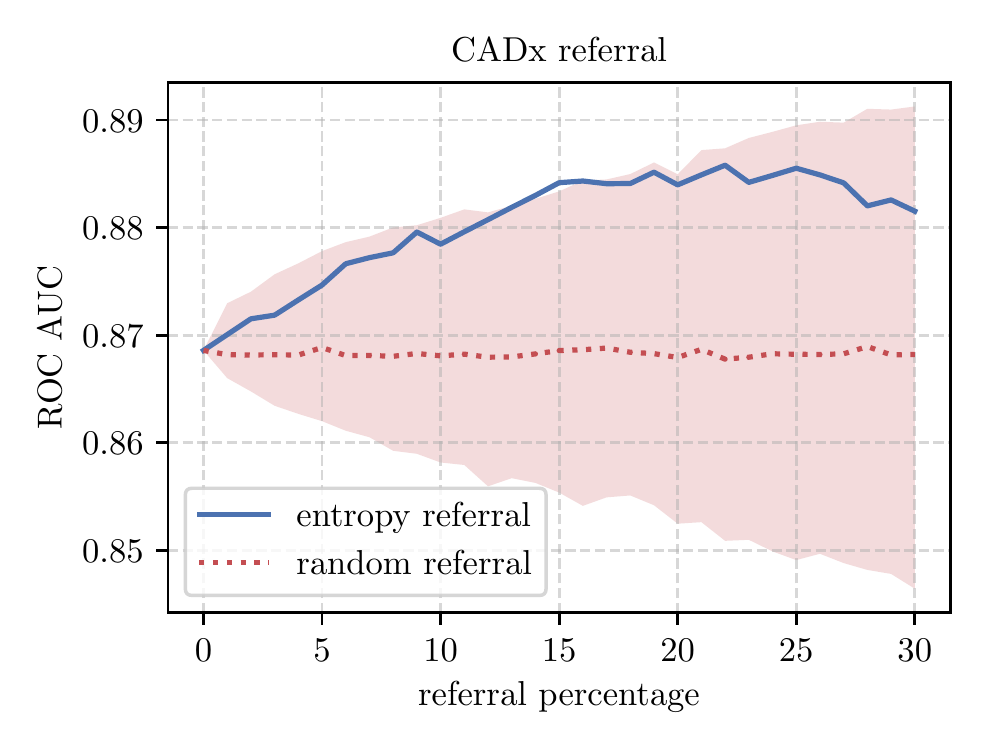}\vspace{-.5cm}
\end{center}
\caption{The CADx area under the ROC curve as a function of referral percentage, for entropy referral strategy and a random strategy with 95\% confidence interval. Note that the confidence interval is wider than the CADe scoring network because the number of patients in the Kaggle test set is smaller than the number of nodules in the LUNA16 dataset.}
\label{fig:cadx_ref}
\end{figure}

\begin{figure}
	\begin{center}
		\includegraphics[width=0.48\textwidth]{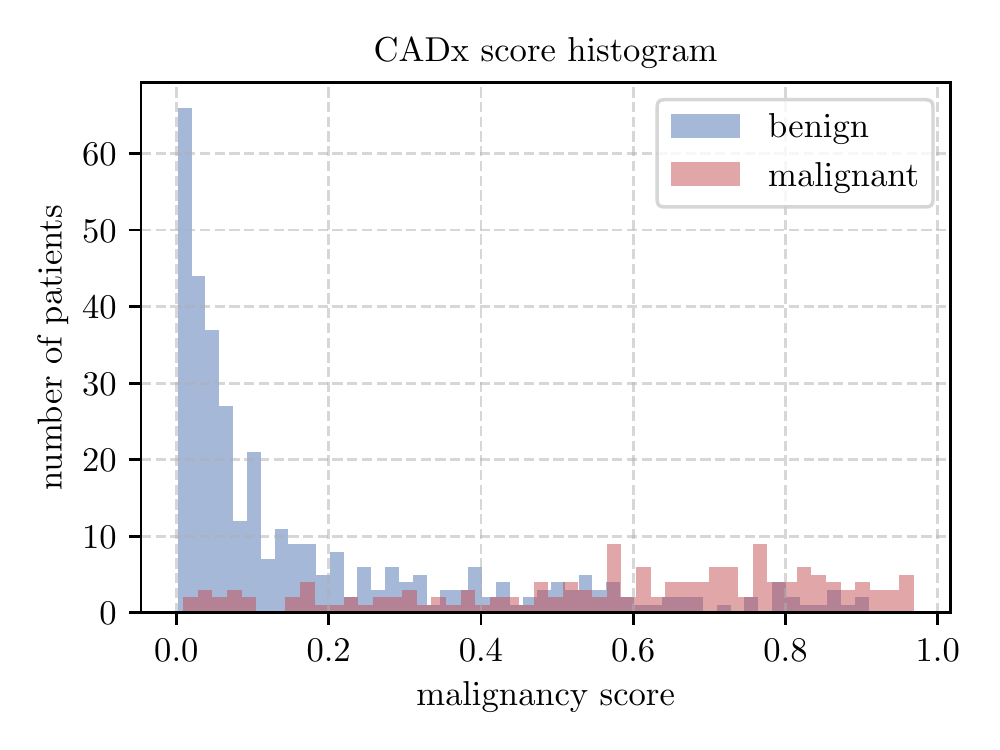}\vspace{-.5cm}
	\end{center}
	\caption{Histogram of CADx malignancy probability estimates for benign and malignant patients. Note that the mode around $0$ for malignant patients is the reason behind false negatives due to missed nodules by the CADe model.}
	\label{fig:cadx_hist}
\end{figure}

Figure \ref{fig:cadx_ref} shows the effect of our referral strategy on the malignancy classification results, where we plot the ROC AUC on the remaining data after referral against different rates of patient referral. We are able to boost the performance on the remaining data and get a maximum AUC of $0.885$, which shows that model uncertainty combined with data uncertainty provide a useful measure to detect patients for which the network is more likely to make an incorrect classification decision. However, the decrease in AUC at higher referral rates indicates that there are incorrect decisions made even when CADx system is very confident. As illustrated by the histogram of CADx malignancy scores in Figure \ref{fig:cadx_hist}, this is a result of patients for whom the CADe system was unable to find a visibly malignant nodule, either because one does not exist, it is too large to be detected by CADe trained on LUNA16 data, or it is simply missed. This is an unavoidable weakness in probability estimations from systems that rely on other systems that are imperfect.

\section{Conclusions}
\label{sec:conc}

In this paper, we introduced a full CADe/CADx system to detect and diagnose lung cancer using low-dose CT scans. Our system uses a cascade of 3D CNNs and achieves state-of-the-art performance on both lung nodule detection and malignancy classification problems on the publicly available LUNA16 and Kaggle datasets. Moreover, we characterized model uncertainty using Monte Carlo dropout and deep ensembles, and showed that quantification of model uncertainty enables our system to provide calibrated classification probabilities, which makes it reliable for subsequent utility/risk-based decision making towards diagnostic interventions or disease treatments. We  demonstrated that we can further improve the performance by using these calibrated probabilities to make patient referral decisions.

Our CADe/CADx system studies demonstrate that CADe and CADx modules should be developed and studied jointly if the goal is to use them as an end-to-end automated diagnostic tool to diagnose lung cancer. This is in contrast to the current design paradigm where CADe and CADx modules are optimized independently for different metrics. We believe that the importance of joint development in radiology is applicable to other systems, where the outputs from radiologists, doctors, and machine learning models are combined to make diagnostic decisions. For example, a machine learning model maximizing the performance on a subset of the dataset that is hard for humans to analyze would be more useful than a model providing the best performance across the whole dataset.

Although our system demonstrates that 3D CNN models can be effectively used to analyze lung CT scans, the performance is still bounded by the limitations of the datasets used to train our models, such as the lack of large nodule annotations in the LUNA16 dataset. We tried to alleviate these problems through intense data augmentation and downsampling, but these solutions are imperfect. An ideal way to train a CADe/CADx would be to have datasets that cover a wider range of nodule sizes and varieties, patients, and scanning equipment. If a malignant lesion is missed by the CADe model, the subsequent CADx model has no way of classifying that scan as malignant, resulting in the biggest source of false negatives in our system. We believe that further investments in curating more inclusive datasets will enable the development of even stronger CADe/CADx models. 

As potential future work, we would like to incorporate patient referral (or reject option) as part the training strategy and learn models that would automatically reject the most uncertain decisions. We would also like to visually analyze learned feature representations to assess whether they could be used as informative biomarkers and help radiologists better understand and interpret CADe/CADx results.

\bibliographystyle{ieeetr}
\bibliography{references} 
\end{document}